\documentclass[journal]{IEEEtai}

\usepackage[colorlinks,urlcolor=blue,linkcolor=blue,citecolor=blue]{hyperref}

\usepackage{color,array}

\usepackage{graphicx}

\usepackage{algorithm}
\usepackage{algorithmic}
\usepackage{amsmath}
\usepackage{booktabs}
\usepackage{multirow} 
\usepackage[table]{xcolor}
\definecolor{cGreen}{RGB}{100,180,100}
\definecolor{cRed}{RGB}{220,50,0}
\definecolor{Klein_Blue}{rgb}{0.0, 0.129, 0.6}
\usepackage{newfloat}
\usepackage{listings}
\usepackage{amssymb}
\usepackage{cite}

\begin{document}

\title{Adaptive Perception for Unified Visual Multi-modal Object Tracking}

\author{Xiantao Hu, Bineng Zhong$^{*}$, Qihua Liang, Zhiyi Mo, Liangtao Shi, Ying Tai, Jian Yang 

\thanks{Xiantao Hu, Bineng Zhong, Qihua Liang and  Liangtao Shi are with the Guangxi Key Lab of Multi-Source Information Mining \& Security, Guangxi Normal University, Guilin 541004, China. }
 \thanks{Zhiyi Mo is currently a Professor in the School of Data Science and Software Engineering, Wuzhou University, Wuzhou 543002, China.}
\thanks{Ying Tai is with the School of  intelligence Science and Technonolgy, Nanjing University, Nanjing 210008, China.}
\thanks{jian Yang is with PCA-Lab, School of Computer Science and Engineering, Nanjing University of Science and Technology, Nanjing 210094,  China.}

}


\maketitle

\begin{abstract}
Recently, many multi-modal trackers prioritize RGB as the dominant modality, treating other modalities as auxiliary, and fine-tuning separately various multi-modal tasks. This imbalance in modality dependence limits the ability of methods to dynamically utilize complementary information from each modality in complex scenarios, making it challenging to fully perceive the advantages of multi-modal. As a result, a unified parameter model often underperforms in various multi-modal tracking tasks. To address this issue, we propose APTrack, a novel unified tracker designed for multi-modal adaptive perception. Unlike previous methods, APTrack explores a unified representation through an equal modeling strategy. This strategy allows the model to dynamically adapt to various modalities and tasks without requiring additional fine-tuning between different tasks. Moreover, our tracker integrates an adaptive modality interaction (AMI) module that efficiently bridges cross-modality interactions by generating learnable tokens. Experiments conducted on five diverse multi-modal datasets (RGBT234, LasHeR, VisEvent, DepthTrack, and VOT-RGBD2022) demonstrate that APTrack not only surpasses existing state-of-the-art unified multi-modal trackers but also outperforms trackers designed for specific multi-modal tasks.
\end{abstract}

\begin{IEEEImpStatement}
Multi-modal tracking is a popular research topic in the field of computer vision. However, most of the current mainstream multi-modal trackers cannot efficiently use the unified parameters of a single model for tracking. By constructing an adaptively perceiving network, my tracker can effectively adapt to the situation where the modal advantages are constantly changing. It improves the accuracy and robustness of multi-modal tracking in complex scenes.
\end{IEEEImpStatement}


\section{Introduction}

\IEEEPARstart{V}{isual} Object Tracking (VOT) is a core task in the field of computer vision, defined as accurately locating any target in subsequent videos using an initial bounding box. It is essential for various tasks~\cite{fang2024guided, lian2023multi, bao2022emotion, anSHaRPoseSparseHighResolution2024, zhang2024few, yan2023dynamic, zheng2023curricular, ning20253d}. Current RGB-based trackers~\cite{aqatrack,evptrack,seqtrack,artrack,FFTrack,siamban,swintrack,CDTrack, xie2024robust} have achieved remarkable results, but inherent limitations in RGB imaging lead to suboptimal performance in challenging and complex scenes.
Therefore, some studies have begun exploring unified multi-modal trackers~\cite{vipt,untrack,onetracker} to fully exploit the complementary strengths of different modalities. These trackers primarily focus on developing a multi-modal framework based on the dominant-auxiliary design principle, aiming to effectively integrate information from various modalities—such as thermal infrared (TIR), event, and depth data into RGB features to enhance tracking performance.

\begin{figure}[t]
  \centering
    \includegraphics[width=1\linewidth,height=4.5cm]
    {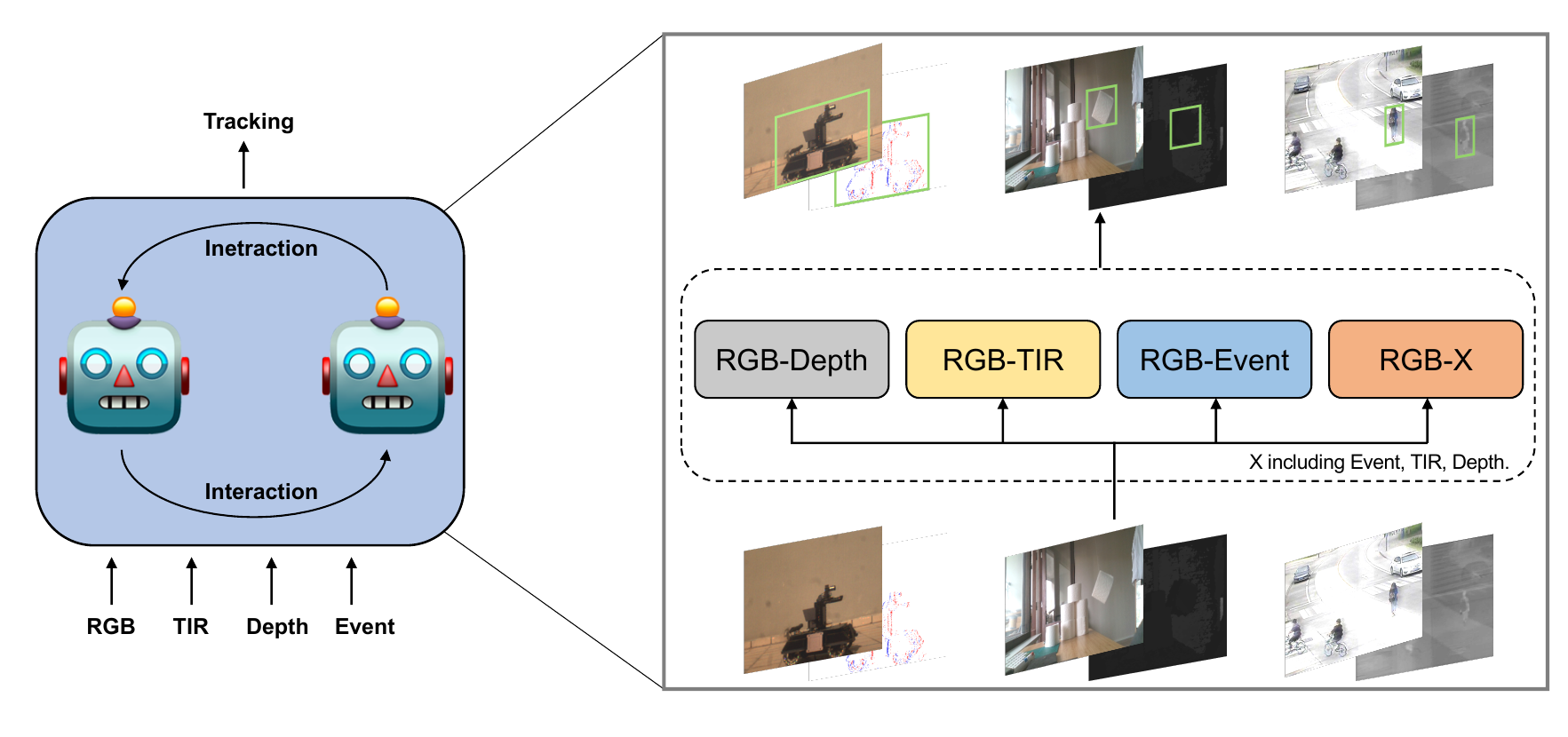}
   \caption{
APTrack is a unified multi-modal tracker that can be applied to various RGB-X tasks (such as RGB-Depth, RGB-TIR, and RGB-Event) with a unified parameters. }
   \label{fig:simple}
\end{figure}

Treating RGB as the dominant modality and others as auxiliary leads the model to overly rely on RGB, neglecting the importance of auxiliary modalities.
This imbalance prevents TIR, event, and depth modalities, which should dominate in challenging scenarios such as low-light conditions, rapid movement, and occlusion, from fully exerting their function.
This imbalance between modalities during the integration of different modal information can lead to reduced perception efficiency, making it more challenging to coordinate and fuse multi-modal features. This challenge hinders the performance of trackers based on unified parameters in handling multi-modal tasks, thereby limiting the model's effectiveness in practical applications and posing an obstacle to achieving a truly unified multi-modal framework.

To address these issues, we propose a unified multi-modal tracker that eliminates the need for a dominant modality. Instead of the traditional dominant-auxiliary approach, our method integrates multiple modalities equally. In other words, APTrack employs the strategy of adaptive perception. It can flexibly adjust the perception and processing based on the inherent features of different modalities and the actual conditions of specific input data. This strategy ensures that each modality is treated equally, allowing them to fully exert their unique characteristics.
For example, in some extremely complex scenarios, the information value contributed by different modalities will change. The adaptive perception mechanism does not rely on fixed weight settings to favor a certain modality, but is able to achieve a balanced utilization of the advantages of each modality.
Furthermore, to further enhance inter-modality correlation and improve modality alignment, we designed an adaptive modality interaction (AMI) module. This module explores a more efficient method of features interaction, enabling effective feature fusion with fewer tokens. Firstly, by learning the learnable tokens for each modality, the model can reduce redundant information and computational load in modality interaction. 
The learning process of these learnable tokens is dynamic, and it can be adjusted adaptively according to the change of input features, without setting fixed markers in advance. Additionally, the global modal perceptor integrates the advantages of each modality by learnable tokens, constructing an effective information flow during the interaction calculation.
This flow is then embedded into the modality features, which enhances modality alignment and facilitates the mutual transmission of advantages between modalities.

In summary, our contributions are as follows:
\begin{itemize}
\item We proposed APTrack, a unified multi-modal tracker that integrates multiple modalities through adaptive perception, eliminating the need for complex multitasking fine-tuning.
\item An adaptive modality interaction module (AMI) is introduced to dynamically adjust interaction features. This enables effective information transmission between modalities using fewer tokens, thereby improving the tracker's performance.
\item The proposed tracker undergoes extensive experimental validation on multiple benchmark datasets. It demonstrates leading performance in uniform training scenarios across multi-modal datasets, showcasing its potential and advantages in the field of multi-modal tracking.
\end{itemize}

\section{Related Works}
\subsection{Unified Multi-modal Tracking}
Recent RGB-based object trackers~\cite{simtrack,mixformer,ostrack,stark,transt} achieved significant advancements in large-scale benchmark tests~\cite{lasot,got10k,trackingnet}. However, the unstable reliability of RGB causes these types of trackers to struggle with complex scenarios such as low-light conditions. Meanwhile, with the development of multi-modal imaging technology and the reduction in costs, an increasing number of multi-modal datasets, such as RGB-T~\cite{lasher,rgbt234}, RGB-E~\cite{visevent}, and RGB-D~\cite{depthtrack,vot-rgbd}, provide possibilities for developing more robust trackers. 
Several works explored how to effectively integrate RGB with another modality. Specifically, TBSI~\cite{tbsi} model used a pair of template features as a mediator to bridge the relationship between TIR search features and RGB search features. BAT~\cite{bat} froze the backbone network and used adapters to transfer features from one modality to another, achieving adaptive fusion between RGB and TIR. STNet~\cite{stnet} combines event images and RGB images from both spatial and temporal perspectives. 
Besides some works have begun exploring unified multi-modal tracking architectures. Among them, ViPT~\cite{vipt} was an early adopter of the dominant-auxiliary modality method to fine-tune multiple multi-modal tracking tasks. OneTracker~\cite{onetracker} utilized large-scale benchmark training of the foundation tracker and incorporated auxiliary branches into the trained model to provide visual cues. On the other hand, Un-Track~\cite{untrack} achieved unified training for multi-modal tasks through low-rank factorization and reconstruction techniques. 
However, these methods still exhibit gaps compared to models specifically designed for single multi-modal~\cite{tbsi,bat,mctrack,hu2024exploiting} tasks.

\begin{figure*}
    \centering
    \includegraphics[height=9.5cm]{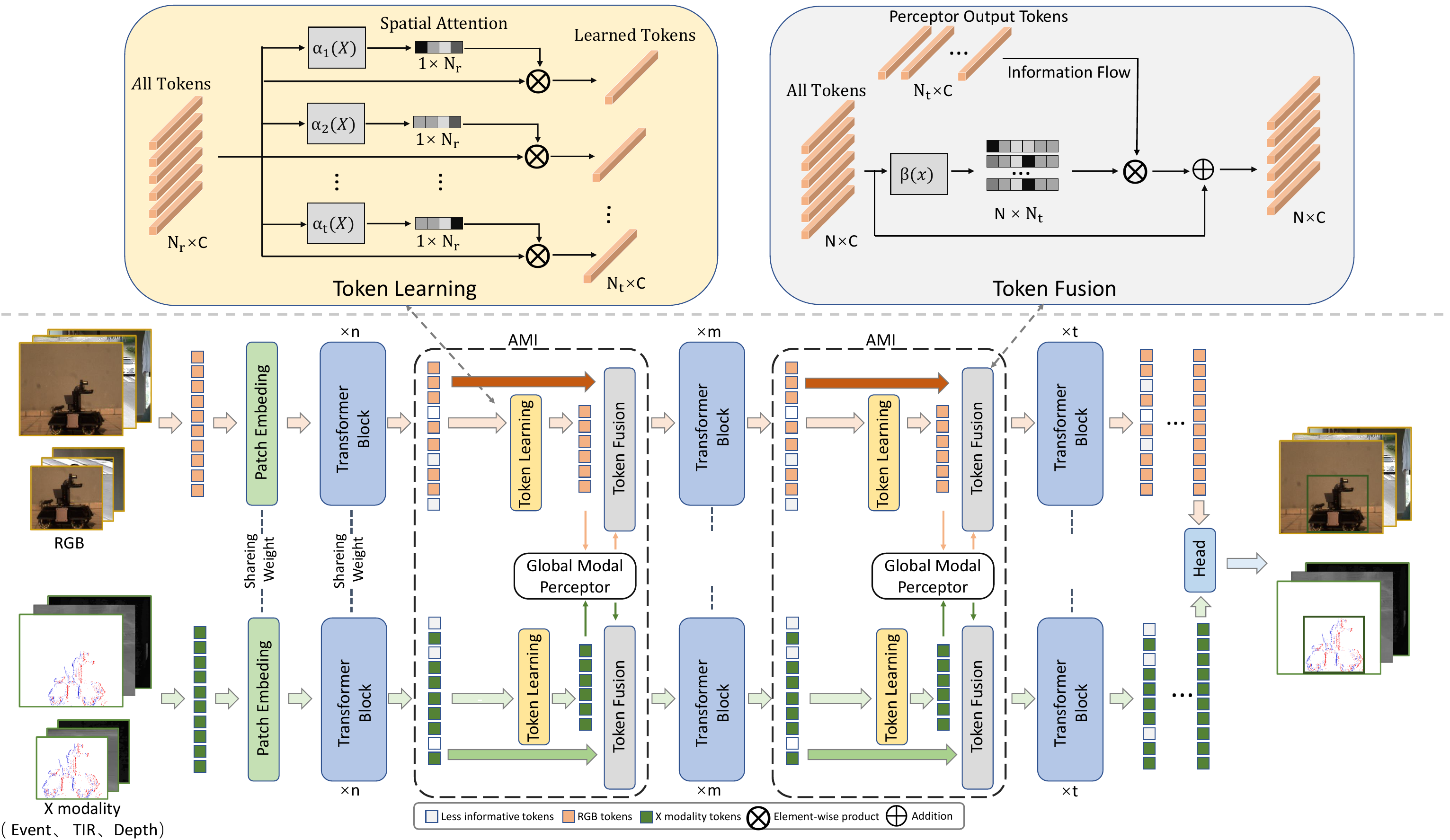}
    \caption{The overall structure of APTrack. APTrack is composed of shared embedding, shared transformer block, AMI and Head. The method of modal processing in this model is completely consistent, and there is no need for extra processing for a certain model, which makes the modal features can be aligned adaptively. In addition, AMI can transfer the advantages of modalities to each other.
 }
  \label{fig:struct} 
\end{figure*}

\subsection{Adaptive Perception}
Recent research~\cite{imagebind,mutex,unified,clip} has shown that through extensive training on vast amounts of datasets, models can acquire the capability to align multiple modalities. However, due to the scarcity of multi-modal tracking datasets, it is not feasible to rely on large-scale data for training.
Therefore, recently unified multi-modal tracker~\cite{vipt,untrack,onetracker,sdstrack} typically rely on the weights of pre-trained RGB-based tracker, optimizing multi-modal task performance by aligning auxiliary modality features with RGB features. However, simply mapping auxiliary modalities to the RGB latent space is insufficient to fully harness their potential, making it challenging to effectively perceive and leverage the advantages of multi-modality. Furthermore, in the modeling part of its shared weights encoder, SDSTrack~\cite{sdstrack} has a bias in the asymmetrical adapter modeling of the X modality and focuses on multitask fine-tuning, while lacking exploration of the application of adaptive perception in multi-modal tasks under uniform parameters. Moreover, research~\cite{learning_li,siamese_av,unified_av}  has shown that using shared-weight local networks for multi-modal feature perception provides significant advantages, particularly in consistent modeling at certain stages. In RGB-T trackers~\cite{tbsi,bat}, this shared feature extraction network has shown a significant performance improvement by virtue of adaptive perception. Based on this observation, we pose a critical question: Can adaptive perception further enhance the performance of unified multi-modal trackers? To explore this, we investigate a strategy that avoids fixing a dominant modality and adopts a consistent modeling approach, aiming to improve the performance of unified multi-modal trackers without the need for separate fine-tuning for each task. In other words, rather than focusing on a single task, our goal is to focus on improving the performance of the unified parameter model to enable its easy deployment in diverse scenarios.

\section{Methodology}
\subsection{Overview}
We introduce a adaptive perception tracking method named APTrack. The overall structure, as illustrated in Fig.~\ref{fig:struct}, shows that APTrack uses an equal modeling strategy, which enables it to quickly and efficiently align features between different modalities, thereby improving the accuracy and stability of multi-modal tracking. Additionally, we introduce an efficient modality interaction module that fully utilizes the advantages of each modality for modality guided. Below, we'll denote modalities other than RGB with the symbol $x$, which could encompass depth, TIR and event data.
\subsection{Adaptive Perception Network}
APTrack employs a consistent modeling approach through the design of adaptive perception to ensure that no individual bias is introduced to any modality. This strategy allows the model to be consistent in processing both input modalities, thus autonomously selecting the best modality based on the input without introducing additional bias.
This approach allows for greater flexibility in adapting to the effectiveness of each modality in different environments.
The input to APTrack comprise two template images $t^{x} \in \mathbb{R}^{3 \times H_t \times W_t}$ and $t^{r} \in \mathbb{R}^{3 \times H_t \times W_t}$ , and two search images  $s^{x} \in \mathbb{R}^{3 \times H_s \times W_s} $ and $s^{r} \in \mathbb{R}^{3 \times H_s \times W_s}$.  
Before the tracking process, the search frames and template frames of the two modalities are tokenized ($t_{x} \in \mathbb{R}^{N_t \times (3 \cdot P^2)}$ , $t_{r} \in \mathbb{R}^{N_t \times (3 \cdot P^2)}$ , $s_{x} \in \mathbb{R}^{N_s \times (3 \cdot P^2)} $ and $s_{r} \in \mathbb{R}^{N_s \times (3 \cdot P^2)}$). Where $P \times P$ is the resolution of each patch. $N_s$ and $N_t$ represent the number of patches for the search and template images, respectively.
Then the same modal data splicing together (that is, $H_ {x} = (t_ {x}, s_ {x}) $ and $H_ {r} = (t_ {r}, s_ {r}) $), and the use of a trainable linear layer with shared weights is employed to project both $H_{x}$ and $H_{r}$ into a D-dimensional space.
Additionally, the same learnable location embedding $P$ is added to generate the RGB tokens $H_{r}^0$ and the X modality tokens $H_{x}^0$. The calculation process is as follows:

    \begin{equation}  H_{r}^0 = [t_{r}^{1}{E};\cdots;t_{r}^{N1}{E};s_{r}^{1}{E};\cdots;s_{r}^{N2}{E}] + p,
    \label{eq:eq2}
    \end{equation}
   \begin{equation}  H_{x}^0 = [t_{x}^{1}{E};\cdots;t_{x}^{N1}{E};s_{x}^{1}{E};\cdots;s_{x}^{N2}{E}] + p ,  \label{eq:eq1} \end{equation}
where ${E} \in \mathbb{R}^{\left(3 \cdot P^{2}\right) \times D}$, $P \in {\mathbb{R}^{N \times D}}$, $N1$ is the number of template tokens and $N2$ is the number of search tokens.

Then $H_{r}^0$  and $H_{x}^0$ will be fed into the transformer block which consists of Multi-head Self-Attention (MSA), LayerNorm (LN), MultiLayer Perceptron (MLP). In this part each layer of the block will also share the weights and process the two modalities separately, the computation process is as follows:

  \begin{equation}
  {h^l} = {H^{l - 1}} + MSA(LN({H^{l - 1}})),
  \end{equation}
  \begin{equation}
    {H^l} = {h^l} + MLP(LN({h^l})),
  \end{equation}
where ${H^{l - 1}} \in [{{H_{r}^{l - 1}},{H_{x}^{l - 1}}}]$ and ${H^{l}} \in [{{H_{r}^{l}},{H_{x}^{l}}}]$. $l$ represent the tansformer block layer number.

In addition, independent modal extraction, lack of communication between modalities, will reduce the multi-modal synergy. So we add adaptive modality interaction module after part of transformer block. The module remains consistent in handling both modes. The purpose of this is to realize the effective transfer of advantages without favoring a certain modality, so that the model can combine the advantages of other models  in the process of interaction.
 \begin{equation}
 {H_r}^l = {H_r}^l + {\rm{AMI(}}H^l_r,H_x^l). \end{equation}
 \begin{equation}
 {H_x}^l = {H_x}^l + {\rm{AMI(}}H^l_x,H_r^l). \end{equation}

\subsection{Adaptive Modality Interaction}
 AMI module mainly includes three stages: token learning, global modal perception and token embedding. Among them, learning token is to reduce the amount of calculation and redundant information, while global modal perceptor is to build an information bridge between modalities. In addition, token embedding is to transfer the information obtained from interaction to modal features. The token learning and
token embedding constitute learnable token mechanism.

\noindent \textbf{Token Learning.} Full-sequence tokens interaction will bring more interaction computation and introduce redundant information. At this stage, the model will learning few tokens through modal features. Specifically, a set of adaptive learnable token is used to significantly reduce the amount of computation in the interaction phase while aggregating features. The model learns the spatial attention weight generated by the input $X$ as the condition, and multiplies it with the $X$ itself. The calculation process is as follows:
 \begin{equation}
F^i =  \gamma ({\alpha _i}({X}))^T \times  X,
\end{equation}
where $F^i$ is the i-th learnable tokens. Linear function $\alpha _i$ is used for intermediate weight tensor computed, $\gamma (\cdot )$ is softmax function. Both modalities will be calculated, and the respective learned token are spliced to generate a new feature sequence $F_{r} \in \mathbb{R}^{N_{t} \times C}$ and $F_{x} \in \mathbb{R}^{N_{t} \times C}$. $ {N_{t}} $ is the number of learnable tokens, and C is the number of channel.

The learnable tokens are used to construct modal interactive information. In this way, subsequent operations require only a small number of tokens to perform efficient modal interaction, significantly reducing the amount of computation.

\begin{figure}[t]
  \centering
    \includegraphics[width=1\linewidth]
    {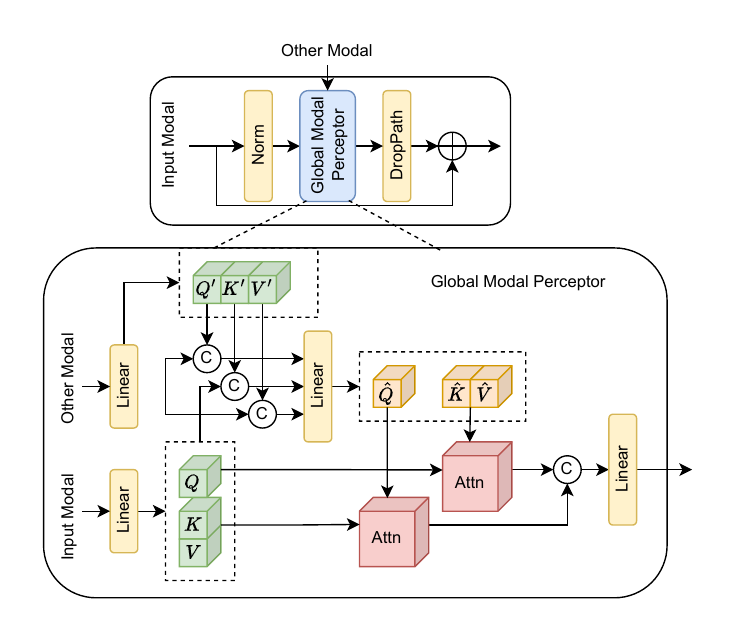}
   \caption{The detailed architecture of Global Modal Perceptor. These perceptors use attention mechanisms to enforce multi-modal global attention. }
   \label{fig:perceptor}
\end{figure}

\noindent \textbf{Global Modal Perceptor.} In this stage, we designed a global modal preceptor for cross-modal interaction, where the learnable tokens serve as the input to the preceptor. As shown in Fig.~\ref{fig:perceptor}, the preceptor includes two core attention mechanisms: (1) Query-guided cross-modal attention (Q-Attention) and (2) Key-Value-guided cross-modal attention (KV-Attention). Within the preceptor, the two modalities interact by alternately serving as the input and other modal. 
 Q-Attention performs cross-modal attention by merging the query vectors of the current modality with those of the other modality. This mechanism generates new query vectors $\hat{Q}$ to focus on important information within the input modal. KV-Attention generates the $\hat{K}$ and $\hat{V}$ vectors by merging the key and value vectors from both modalities, thereby facilitating the retrieval of complementary information between them. The process can be described as follows:
 
 \begin{equation}
\begin{gathered}
W^Q = Attention(\hat{Q},K,V), \\
W^{KV} = Attention(Q,\hat{K},\hat{V})
  \end{gathered}
\end{equation}

Subsequently, $W^Q$ and $ W^{KV}$ are fused through a linear layer and fed back into the input learnable tokens, enhancing the information flow with augmented modal information. 
Importantly, this process is reciprocal, with each modality taking turns as the input modality for independent computation.
\begin{table*}[t]\footnotesize
  \centering
    \caption{State-of-the-art comparison on LasHeR~\cite{lasher} (RGB-T), VisEvent~\cite{visevent} (RGB-E), and DepthTrack~\cite{depthtrack} (RGB-D) benchmarks. We use \textcolor{gray}{gray} color to denote our trackers. The best three real-time results are shown in \textbf{\textcolor{cRed}{red}}, \textcolor{blue}{blue} and \textcolor{cGreen}{green} fonts.}
  \setlength{\tabcolsep}{3mm}{  
  \resizebox{0.9\textwidth}{!}{
  \begin{tabular}{c|c|c|c cc c cc c ccc}
    \toprule
    & \multirow{2}*{Method} & \multirow{2}*{Source} &  \multicolumn{2}{c}{LasHeR} & & \multicolumn{2}{c}{VisEvent} & &   \multicolumn{3}{c}{DepthTrack} \\
    \cline{4-5}
    \cline{7-8}
    \cline{10-12}
   & & & Suc. &Pr. & &Suc. &Pre.  &  & F-score&Re. &Pr. \\
    \midrule[0.5pt]
    \multirow{7}*{\rotatebox{90}{Unified}}
     & \cellcolor{gray!15}APTrack &\cellcolor{gray!15}Ours &\cellcolor{gray!15}\textcolor{cRed}{58.9} &\cellcolor{gray!15}\textcolor{cRed}{74.1} &\cellcolor{gray!15} &\cellcolor{gray!15}\textcolor{cRed}{61.8} &\cellcolor{gray!15}\textcolor{blue}{78.5} &\cellcolor{gray!15} &\cellcolor{gray!15}\textcolor{cRed}{62.1} &\cellcolor{gray!15}\textcolor{cRed}{61.9} &\cellcolor{gray!15}\textcolor{cRed}{62.3}\\
     & Un-Track~\cite{untrack} &CVPR24 &51.3 &64.6 & &58.9 &75.5 & &\textcolor{cGreen}{61.0} &\textcolor{blue}{61.0} &\textcolor{cGreen}{61.0}\\
     & SeqTrack~\cite{seqtrack} &CVPR23 &49.0 &60.8 & &50.4 &66.5 & &59.0 &60.0 &58.0\\
     & ViPT~\cite{vipt} &CVPR23 &49.0 &60.8& &57.9 &74.0 & &56.1 &56.2 &56.0\\
     & OSTrack~\cite{ostrack} &ECCV22 &42.2 &53.0 & &52.5 &69.1 & &56.9 &58.2 &55.7\\
     & AiATrack~\cite{aiatrack} &ECCV22  &36.5 &46.3 & &44.4 &62.6 & &51.5 &52.6 &50.5\\
     & Stark~\cite{stark} &ICCV21 &33.3 &41.8 & &52.5 &44.8 & &39.7 &40.6 &38.8\\
     \midrule[0.5pt]
      \multirow{7}*{\rotatebox{90}{Specific}}
      & AINet~\cite{AINet} &AAAI25 &\textcolor{blue}{58.2} &\textcolor{blue}{73.0} & &- &- & &- &- &- \\
      & CAFormer~\cite{CAFormer} &AAAI25 &55.6 &70.0 & &- &- & &- &- &- \\
    & OneTracker~\cite{onetracker} &CVPR24 &53.8 &66.5 & &\textcolor{blue}{60.8} &\textcolor{cRed}{78.6} & &60.9 &60.4 &60.7\\
     & SDSTrack~\cite{sdstrack} &CVPR24 &53.1 &66.7 & &\textcolor{cGreen}{59.7} &\textcolor{cGreen}{76.7} & &\textcolor{blue}{61.4} &\textcolor{cGreen}{60.9} &\textcolor{blue}{61.9}\\
     & BAT~\cite{bat} &AAAI24 &56.3 &70.2 & &- &- & &- &- &-\\
     & GMMT~\cite{gmmt} &AAAI24 &\textcolor{cGreen}{56.6} &\textcolor{cGreen}{70.7} & &- &- & &- &- &-\\
     & Depthrefiner ~\cite{dualrgbd_depthrefiner}  &ICME24 &- &- & &- &- & &51.0 &50.7 &51.3 \\
     & MMHT~\cite{dualevent_mmht} &Arxiv24 &- &- & &55.3 &73.4 & &- &- &- \\
     & ViPT~\cite{vipt} &  CVPR23 &52.5 &65.1 & &59.2 &75.8 & &59.4 &59.6 &59.2\\
     & TBSI~\cite{tbsi} &CVPR23 &56.3 &70.5 & &- &- & &- &- &-\\
     & ProTrack~\cite{protrack} &  MM22 &42.0 &53.8 & &47.1 &63.2 & &57.8 &57.3 &58.3\\
\bottomrule

\end{tabular}
  
}

  \label{tab:rgb-x}
}
\vspace{-2.5mm}
\end{table*}

   \begin{table*}[t]
    \centering
    \caption{Overall performance on VOT-RGBD2022~\cite{vot-rgbd}. The best three results are highlighted in {\color{cRed}red}, {\color{blue}blue}, {\color{cGreen}green}, respectively.}
    \resizebox{1\linewidth}{!}{
    \begin{tabular}{c|cccccc|cccccc}
     \toprule
    & \multicolumn{6}{c|}{Specific Parameters} & \multicolumn{6}{c}{Unified Parameters}\\
    \toprule
       \multirow{2}*{Method} &DeT &SPT &ProTrack &ViPT  &OneTracker &SDSTrack &Stark &AiATrack  &OSTrack  &SeqTrack  &Un-Track   &APTrack \\
       &\cite{det} &\cite{spt} &\cite{protrack} &\cite{vipt} &\cite{onetracker} &\cite{sdstrack} &\cite{stark} &\cite{aiatrack} &\cite{ostrack} &\cite{seqtrack} &\cite{untrack} &\textbf{(Ours)}
       \\ 
       \toprule
       EAO($\uparrow$) &65.7 &65.1 &65.1 &72.1 &{\color{cGreen}72.7} &{\color{blue}72.8} &44.5 &64.1
       &66.6 &67.9 &71.8 &{\color{cRed}77.4}\\ 
       Accuracy($\uparrow$)&76.0 & 79.8 &80.1 &81.5&{\color{cGreen}81.9}&81.2 &71.4 &76.9
       
       &80.8&80.2 &{\color{blue}82.0} &{\color{cRed}82.1}\\  
        Robustness($\uparrow$) &84.5 &85.1 &80.2 &87.1 &{\color{cGreen}87.2}&{\color{blue}88.3}
        &59.8 &83.2
        
        &81.4 &84.6 &86.4 &{\color{cRed}93.4}\\
    \bottomrule
    \end{tabular}
    }        
    \label{tab:vot-rgbd2022}
    \end{table*}

\noindent \textbf{Token Embedding.} Modal interaction is essentially a process of mutual notification. One modality will inform another modality of its own advantageous information about the target, and the other modality will integrate the advantageous information with its own mined features. After we use the learnable tokens to generate interactive information, we need to embed the information into the input features of the AMI module. We use $F \in \mathbb{R}^{{N_t} \times C}$  to refer to the interactive information of ones modal, and $H^{l+1} \in \mathbb{R}^{{N} \times C}$ to represent the feature sequence with N number of token input into the AMI. Specifically, we embed learnable tokens into $H^{l}$ by learning to combine tokens at each spatial location.
\begin{equation}
{H^{l + 1}} = {B_w}F + {H^{l}},
 \end{equation}
where ${B_w \in \mathbb{R}^{N_t \times N_t}}$ is implemented through simple linear layer and a softmax function, it is the spatial weight tensor.

\subsection{Prediction Head and Loss Function}
\noindent \textbf{Prediction Head.} Because of the adaptive perception modeling method and effective modal interaction strategy, the model can accurately extract the multi-modal alignment feature.
So it is only necessary to simply splice the two modal features, and accurate target positioning can be realized in the prediction head. 
Specifically, we employ conventional classification heads and bounding box regression. The classification score map, denoted by $ P \in \mathbb{R}^{ \frac{H_{s}}{P} \times \frac{W_{s}}{P}}$, the local offset, symbolized as $  O \in \mathbb{R}^{2 \times \frac{H_{s}}{P} \times \frac{W_{s}}{P}}$, and the bounding box size, represented by $  S \in \mathbb{R}^{ 2 \times \frac{H_{s}}{P} \times \frac{W_{s}}{P}}$, are all obtained by the prediction modules, respectively. \
Obtain accurate tracking through rough localization, offset, and box size,as
 \begin{equation} 
    \begin{aligned}
     & x = x_{d} + O(0,x_{d},y_{d}), \\
     & y = y_{d} + O(1,x_{d},y_{d}), \\
     & w = S(0,x_{d},y_{d}), \\
     & h = S(1,x_{d},y_{d}), \\
    \end{aligned}
    \label{eq:bbox}
     \end{equation}
where $(x_{d},y_{d})$ denotes the coordinates which is the highest
score of classification score map.

\noindent \textbf{Loss Function.} We select focal loss~\cite{focal_loss} as our classification loss, $L_{cls}$, while assigning the $L_1$ loss and GIoU loss~\cite{giou} as our regression loss types. Consequently, the comprehensive loss L is determined as follows:
    \begin{equation}
    L = L_{cls} + \lambda_{1}L_{1} + \lambda_{2}L_{GIoU},
    \end{equation}
where $\lambda_{1}=5 $ and  $\lambda_{2}=2 $ are the regularization parameters.

\begin{figure*}[t]
  \centering
    \includegraphics[width=1\linewidth]
    {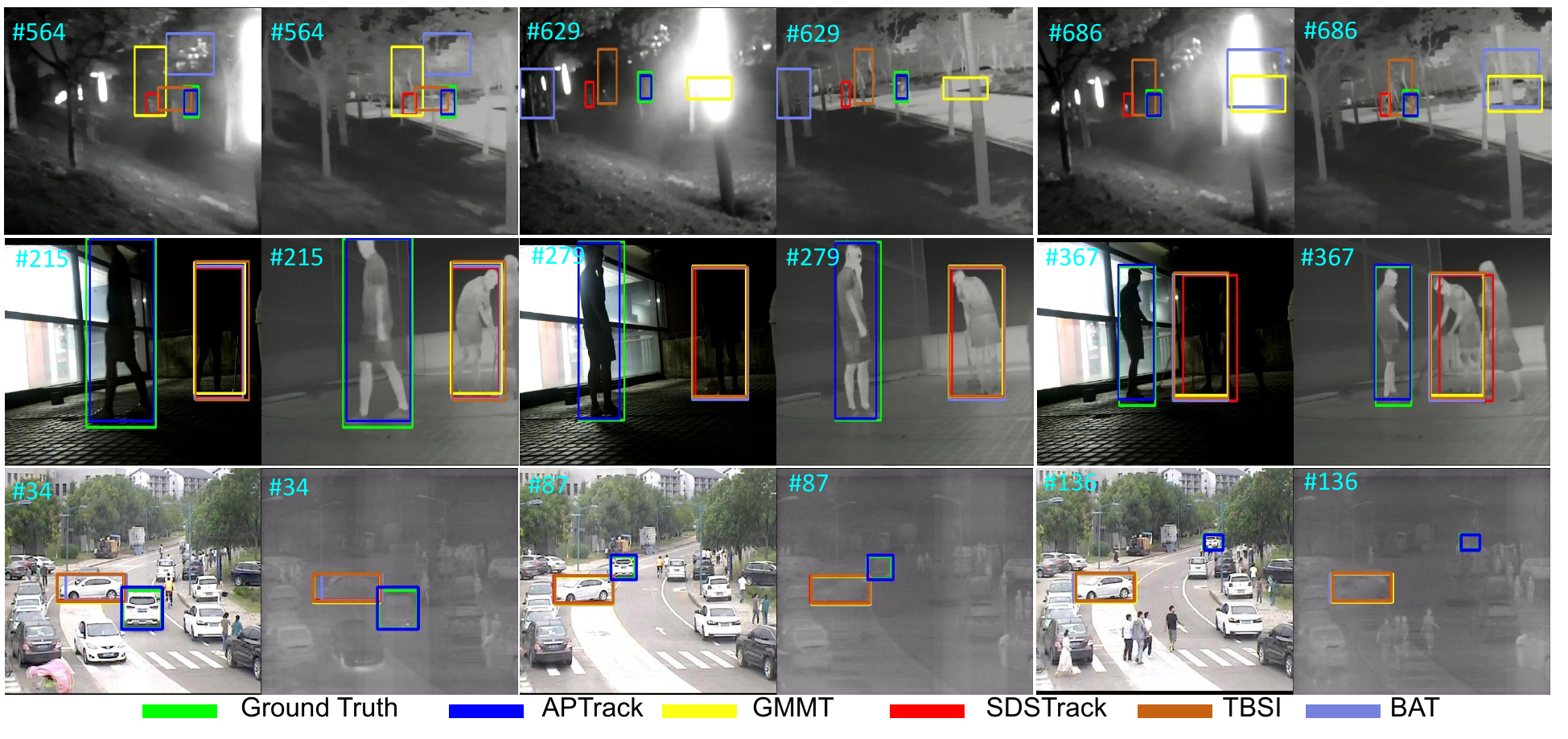}
   \caption{Visual result of RGB-T. The three sequences from top to bottom represent the following scenarios: both modalities exhibit low effectiveness, the RGB modality is less effective than the TIR modality, and the TIR modality is less effective than the RGB modality. }
   \label{fig:tracker_contrast}
\end{figure*}

\begin{figure}
  \centering
    \includegraphics[width=1\linewidth,height=5cm]
    {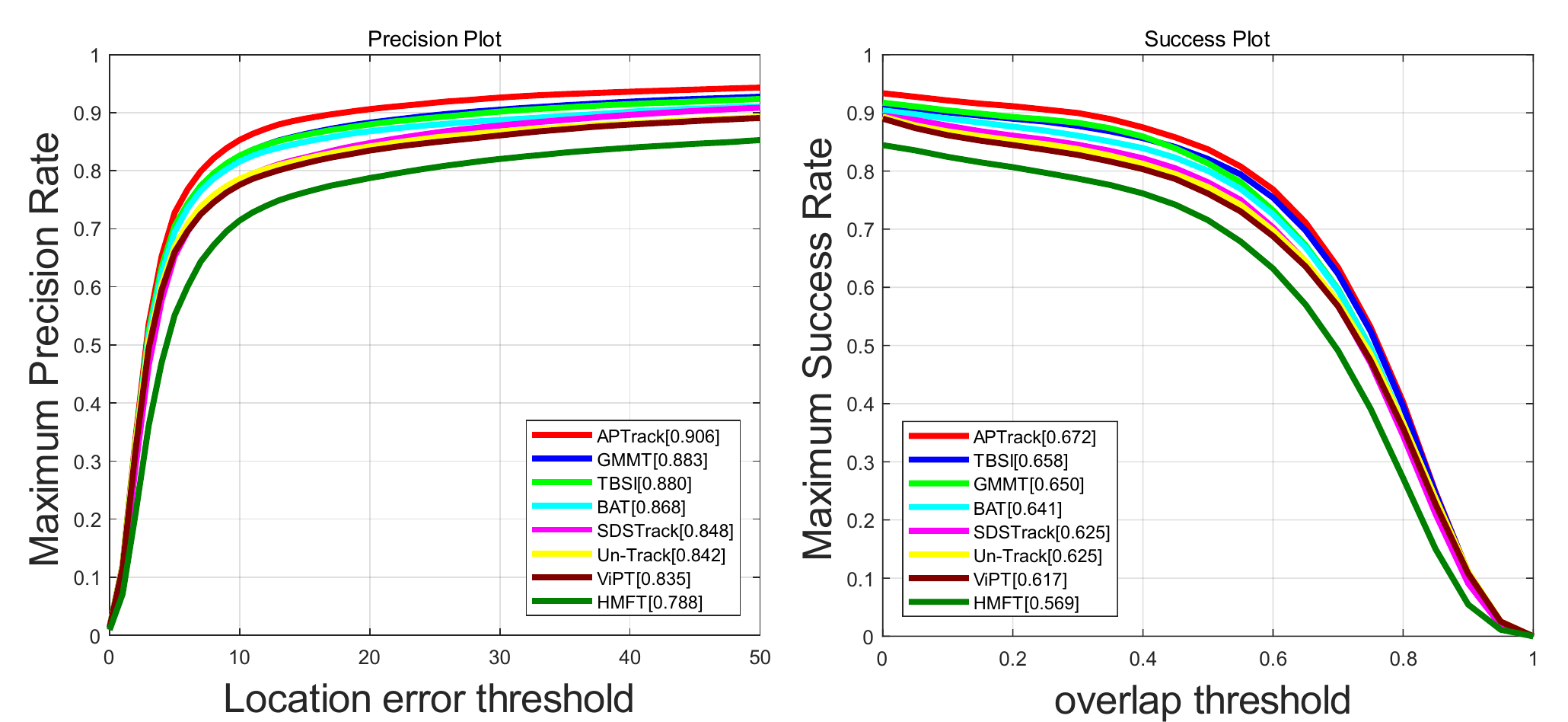}
   \caption{More MPR/MSR comparisons on RGBT234. }
   \label{fig:rgbt234}
\end{figure}

\section{Experiment}
\subsection{Implementation Details}

To establish a truly unified multi-modal tracking framework, we developed a versatile RGB-X tracker capable of flexibly addressing a range of multi-modal tasks, including RGB-T, RGB-D, and RGB-E tracking. The training process incorporated the VisEvent~\cite{visevent}, DepthTrack~\cite{depthtrack}, and LasHeR~\cite{lasher} datasets. We conducted the training on four NVIDIA GeForce RTX 4090 GPUs, running for 20 epochs. Each epoch comprised 60,000 sample pairs with a batch size of 16, using AdamW~\cite{adamw} as the optimizer. The initial learning rate was set at 1e-4 for the AMI module and 1e-5 for other parameters. After 10 epochs, the learning rate was reduced by a factor of 10. It is worth noting that our tracker employs a dual-template mechanism, which consists of an initial template frame and a dynamic template frame. By default, the dynamic template will be updated when the update interval reaches 5 and the classification score exceeds 0.65. Specifically, the initial template frame remains constant during the tracking process, while the dynamic template frame is updated according to the confidence level of the tracking results.

 We tested our model on an NVIDIA GeForce RTX 4090, achieving a runtime speed of 50.5 FPS. A single model with unified parameters was used for multitasking evaluation. For the RGB-T task, we tested on LasHeR~\cite{lasher} and RGBT234~\cite{rgbt234}. The RGB-E task was evaluated using VisEvent~\cite{visevent}, while the RGB-D task was tested on DepthTrack~\cite{depthtrack} and VOT-RGBD2022~\cite{vot-rgbd}. For comparison purposes, we categorize the models into two types: Specific parameter models that require fine-tuning for specific tasks, and unified parameter models that do not require any fine-tuning.

\subsection{Comparison with State-of-the-arts}

\noindent \textbf{LasHeR}. LasHeR~\cite{lasher} benchmark is a large RGB-T tracking dataset containing 1224 aligned sequences with a total of over 2$\times$730K frames, caputered in various types of imaging platforms.
This dataset covers a wide range of object categories, camera perspectives, scene complexities, and environmental factors, while taking into account various real-world challenges.  We compared with recent high-performance multi-modal trackers, including AINet~\cite{AINet}, CAFormer~\cite{CAFormer}, GMMT~\cite{gmmt}, BAT~\cite{bat}, TBSI~\cite{tbsi}, ViPT~\cite{vipt}, OneTracker~\cite{onetracker}, SDSTrack~\cite{sdstrack}, ProTrack~\cite{protrack} and Un-Track~\cite{untrack}. It is worth noting that the GMMT~\cite{gmmt}, BAT~\cite{bat} and TBSI~\cite{tbsi} is specifically designed for RGB-T tracking tasks.
As shown in Tab.~\ref{tab:rgb-x}, APTrack outperformed AINet~\cite{AINet} in both precision and success, achieving 1.1\% and 0.7\% improvements, respectively. The results from the large-scale RGB-T dataset LasHeR~\cite{lasher} not only verifies the validity of our model, but also fully demonstrates that adaptive perception modeling method and effective modal interaction module can significantly boost performance.  

\noindent \textbf{VisEvent.} VisEvent~\cite{visevent} contains 500 training video sequences and 320 testing video sequences, which is currently the largest RGB-E dataset. In VisEvent~\cite{visevent}, we compared with recent high-performance multi-modal trackers, including ViPT~\cite{vipt}, SDSTrack~\cite{sdstrack}, ProTrack~\cite{protrack} OneTracker~\cite{onetracker}, Depthrefiner~\cite{dualrgbd_depthrefiner} and Un-Track~\cite{untrack}.
As shown in Tab.~\ref{tab:rgb-x}, under the same strategy of unified parameters model, our tracker exceeds Un-Track~\cite{untrack} by 2.9\% in success. This is due to the fact that our network does not have a fixed pattern of dominant-auxiliary modes, allowing the model to adaptively make choices of modes in complex changing dynamic scenarios.

\noindent \textbf{DepthTrack.} DepthTrack~\cite{depthtrack} is a long-time tracking dataset in which the average sequence length is 1473 frames. The dataset covers 200 sequences, 40 scenes and 90 target objects.
In Depthtrack~\cite{depthtrack}, precision (Pr), recall (Re), and F-score are used to measure the tracker performances. where F-score, calculated by $f = \frac{2RePr}{Re+Pr}$, is the primary measure. In DepthTrack~\cite{depthtrack}, we compared with recent high-performance multi-modal trackers, including SDSTrack~\cite{sdstrack},  ViPT~\cite{vipt}, ProTrack~\cite{protrack} OneTracker~\cite{onetracker}, MMHT~\cite{dualevent_mmht}  and Un-Track~\cite{untrack}.
As shown in Tab.~\ref{tab:rgb-x}, compared to SDSTrack~\cite{sdstrack}, APTrack benefits from the ability to adaptively gain modal advantage in complex and variable dynamic scenes, achieving a 0.7

 \noindent \textbf{RGBT234.} The RGBT234~\cite{rgbt234} benchmark extends the RGBT210~\cite{rgbt210} dataset, providing enriched annotations and a wider array of environmental challenges.
It comprises a total of 234 sequences, consisting of 234K highly aligned pairs of RGBT videos with a maximum of 8K frames per sequence, and offers 12 attributes. 
Due to inconsistencies between the ground truths of RGB and TIR in this dataset, the evaluation methodology of RGBT210 is discarded. Instead, the benchmark employs the Maximum Precision Rate (MPR) and Maximum Success Rate (MSR) as assessment metrics, offering a more comprehensive evaluation of tracker performance. 
Specifically, for each frame, accuracy is determined by calculating the Euclidean distance between the predicted bounding box and the ground truth separately in RGB and TIR modalities, selecting the smaller distance for computation. 
As shown in Fig.~\ref{fig:rgbt234}, in the RGB-T domain, only APTrack outperforms the models specifically designed for RGB-T among the unified models. Specifically, our tracker outperformed GMMT~\cite{gmmt} by 1.4\% in MSR.

 \noindent \textbf{VOT-RGBD2022}. VOT-RGBD2022~\cite{vot-rgbd} consists of 127 short RGB-D sequences. It employs Accuracy, Robustness, and Expected Average Overlap (EAO) as evaluation metrics. As shown in Tab.~\ref{tab:vot-rgbd2022}, APTrack exceeds the SDStrack~\cite{sdstrack} by 4.6\% in EAO.





\begin{table}
  \centering
\caption{Ablation studies on the effect of various components on LasHeR~\cite{lasher} testing set. AP, GMP, LT denote adaptive perception structure, global modal perceptor, and learnable token mechanism (include token learning and token embedding). We use \textcolor{gray}{gray} color to denote our final trackers setting.}
 \resizebox{1\linewidth}{!}{
  \setlength{\tabcolsep}{2mm}{
  \begin{tabular}{c|cccc|cc}
\toprule
Variants &Baseline &AP  & GMP  & LT &Precision &Success \\
\midrule
1 &\checkmark  &- &- &- &65.7 &52.3 \\
2 &\checkmark  & \checkmark   &- &- &71.3 &56.9 \\
3 &-   &\checkmark & \checkmark  &- &72.9 &58.7\\
4 &-  &\checkmark  & -  &\checkmark  &73.1 &58.1 \\
\cellcolor{gray!15}5 &\cellcolor{gray!15}-  &\cellcolor{gray!15}\checkmark  & \cellcolor{gray!15}\checkmark  &\cellcolor{gray!15}\checkmark  &\cellcolor{gray!15}74.1 &\cellcolor{gray!15}58.9 \\
\midrule
\end{tabular}
}
}

\label{tab:ablation_components}
\end{table}

\begin{table}
  \centering
\caption{Ablation studies about number of learnable tokens. We use \textcolor{gray}{gray} color to denote our final trackers setting. }
 \resizebox{0.6\linewidth}{!}{
  \setlength{\tabcolsep}{2mm}{
  \begin{tabular}{c|cc}
\toprule
  Number   &Precision &Success \\

\midrule
 0 &72.9 &58.7\\
16 &73.2 &58.3\\
\cellcolor{gray!15}32 &\cellcolor{gray!15}74.1 &\cellcolor{gray!15}58.9\\
64 &73.9 &58.6 \\ 
\midrule
\end{tabular}
}}

\label{tab:number}
\end{table}

\begin{table}
  \centering
  \caption{Comparison of inference speed with sota modal. All methods are tested in an environment identical to our APTrack. }
  \resizebox{0.75\linewidth}{!}{
  \begin{tabular}{c|c|c|cc}
\toprule
Methond  &FPS  &Precision &Success \\

\midrule
ViPT~\cite{vipt} &41.5  &65.1 &52.5\\
SDSTrack~\cite{sdstrack} &23.5 &66.5 &53.1\\
BAT~\cite{bat} &47.8 &70.2 &56.3\\
TBSI~\cite{tbsi} &48.0 &70.2 &56.5\\
\midrule
APTrack (ours) &\textbf{50.5} &\textbf{74.1} &\textbf{58.9}\\
\midrule
\end{tabular}
}

\label{tab:fps}
\end{table}

\begin{figure*}[t]
  \centering
    \includegraphics[width=1\linewidth,height=10.5cm]
    {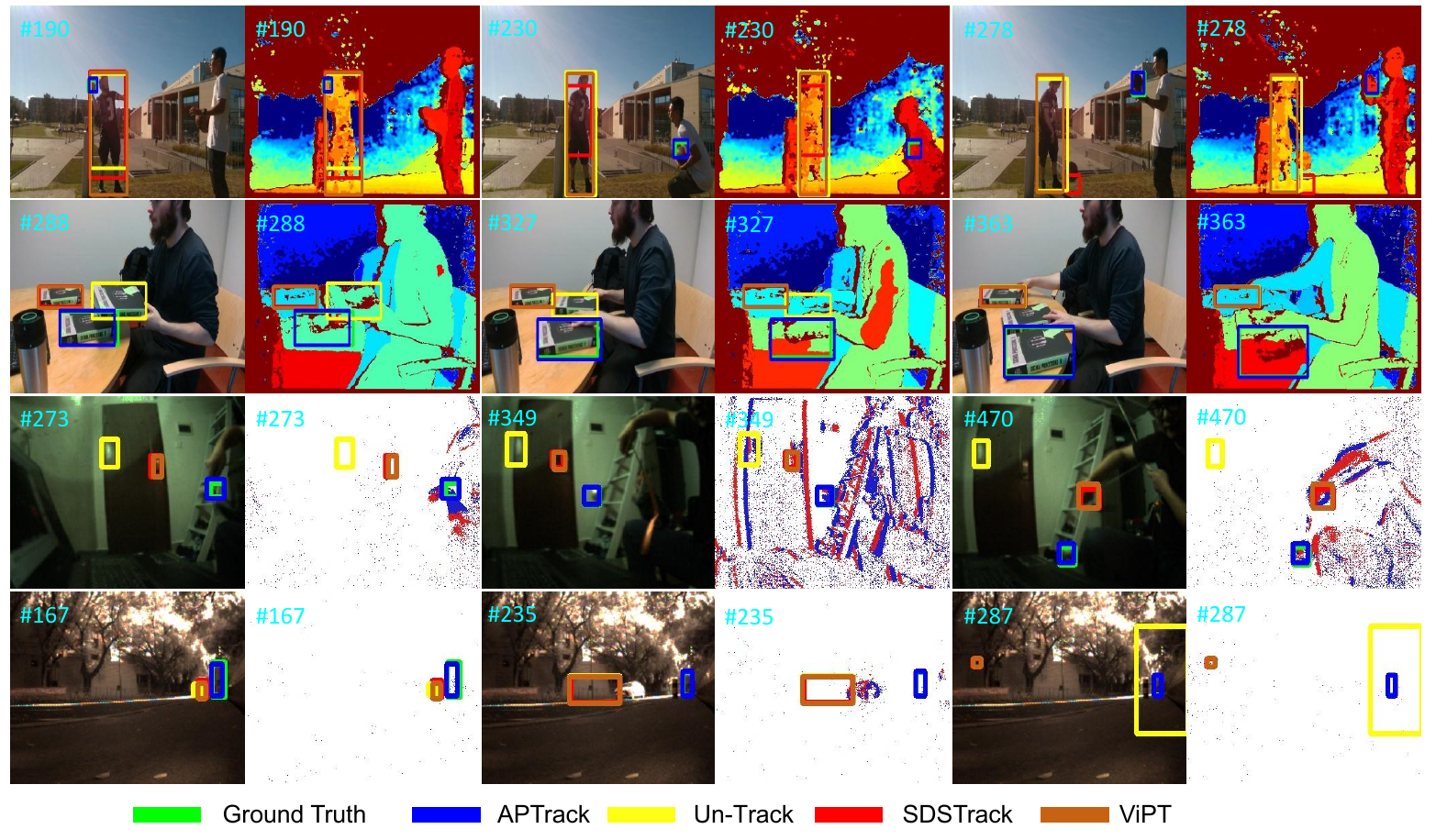}
   \caption{Visualization comparison of our method with other unified multi-modal trackers on RGB-E and RGB-D tasks. Four challenging scenarios including occlusion, similar object interference, cluttered backgrounds and small targets are shown in from top to bottom. }
   \label{fig:contrast}
\end{figure*}


\subsection{Exploration Study and Analysis.}
Our model exhibits a unified characteristic, meaning that the same set of parameters can be adapted to multiple tasks. In this section, we refer to Un-Track~\cite{untrack} to perform ablation experiments on a single multi-modal task, specifically we will perform ablation experiments and explorations on the LasHeR~\cite{lasher} test set to validate the model components.

\noindent \textbf{Components Analysis.} 
We used the dual-template ViPT~\cite{vipt} as our baseline.
As summarized in Tab.~\ref{tab:ablation_components}, the adaptive perception structure achieved a 4.3\% improvement in success rate due to the absence of bias towards any modality. This improvement is attributed to the fact that in complex environments, the effectiveness of each modality is not constant. The model, through self-selection, can flexibly adapt to the varying importance of different modalities, thereby enhancing overall tracking performance. Moreover, the cross-modal information flow generated by our global modal preceptor significantly enhances performance. Additionally, the learnable token mechanism, by learning more representative features, synergizes with the preceptor, further boosting the model's accuracy.


\begin{figure*}
  \centering
    \includegraphics[width=0.85\linewidth,height=6.5cm]
    {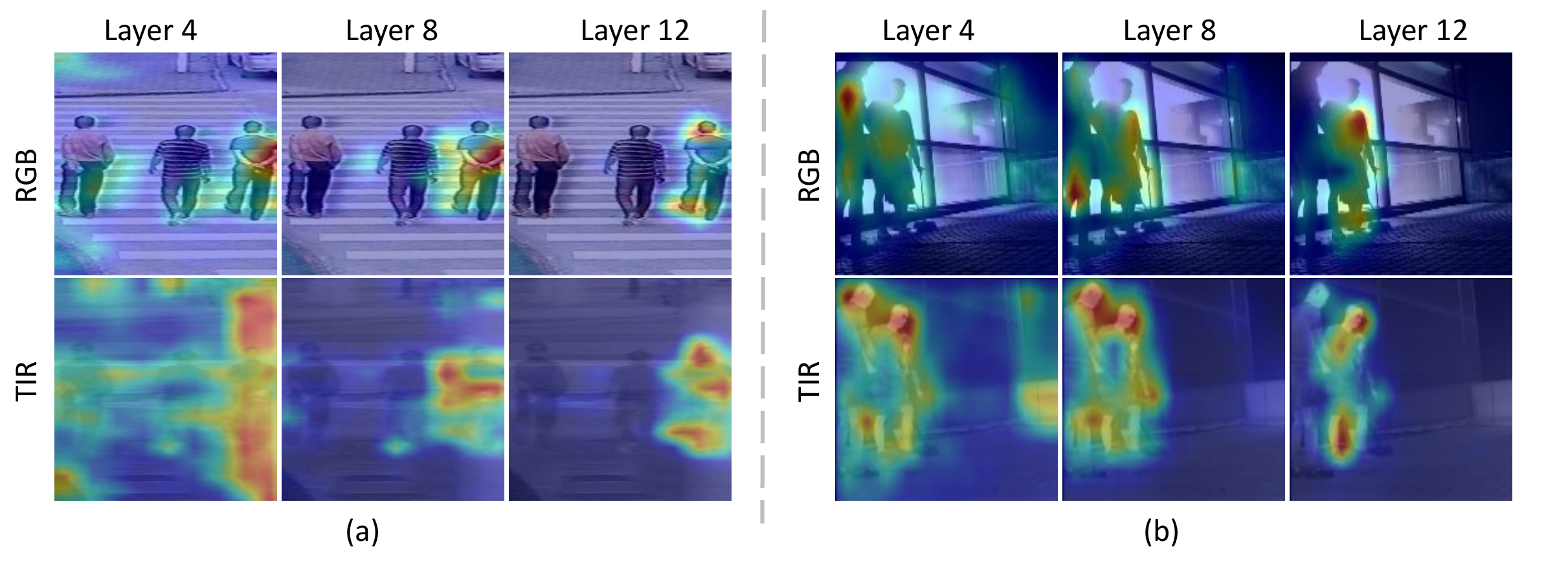}
   \caption{Visualization of the attention maps for a representative pair sequence. The modal dominance of these two sequences is not consistent, specifically (a) the sequence with superior RGB imaging has better target features and (b) it is the TIR that has better target features.}
   \label{fig:heat}
\end{figure*}

\noindent \textbf{Number of learnable Tokens Analysis.}
We analyze the ratio of learnable token to the original token.
Learning 0 tokens indicates that no learnable token mechanism is applied, and the model directly performs feature interaction. As shown in the Tab.~\ref{tab:number}, even a small number of learnable tokens can enhance performance, as reducing the number of tokens helps aggregate information, minimize redundancy, and avoid unnecessary interactions. However, if the number of learnable tokens is too small, it may not adequately capture the target's features, thus limiting performance improvements.


\begin{figure}
  \centering
    \includegraphics[width=0.9\linewidth]
    {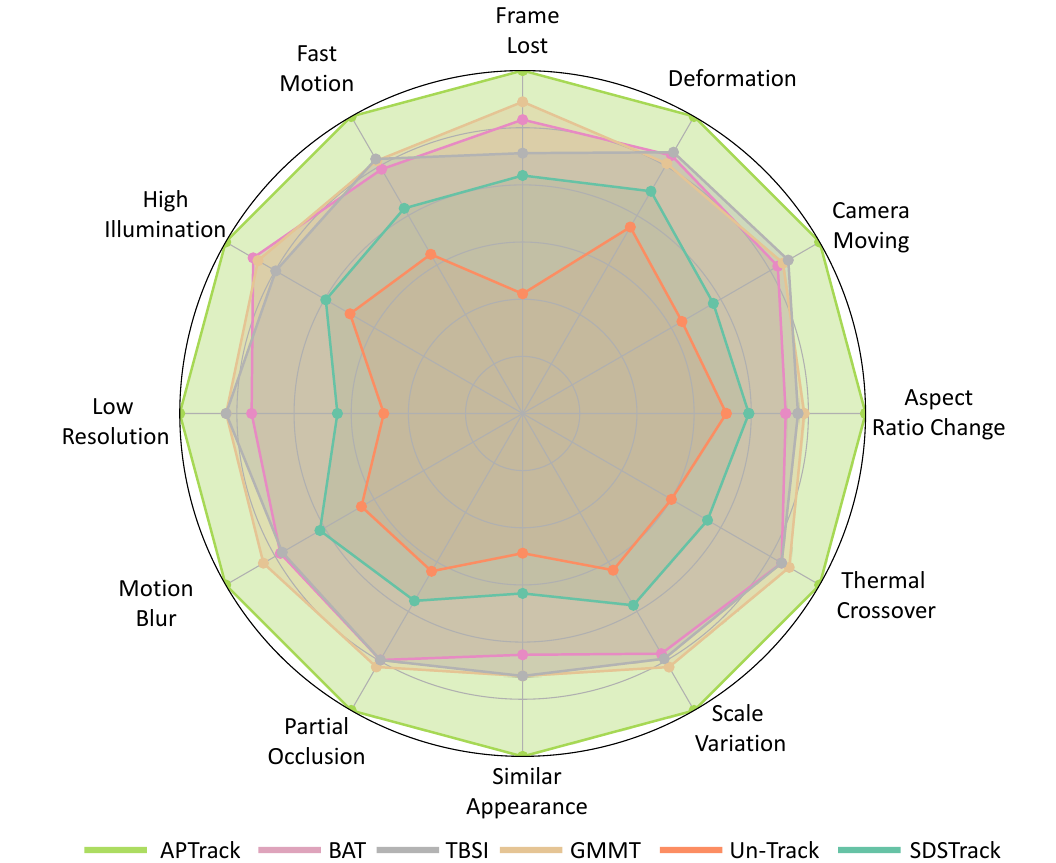}
   \caption{Success rate of different attributes on the LasHeR~\cite{lasher} test set. }
   \label{fig:attribute}
\end{figure}


\noindent \textbf{Qualitative Comparison.} In order to comprehensively analyze the performance of our method in various scenarios, we performed attribute evaluation as well as visualization on the LasHeR~\cite{lasher} dataset.
As shown in Fig.~\ref{fig:tracker_contrast}, with different cases of modal validity, our model can track the target better. And, in Fig.~\ref{fig:attribute},  APTrack can surpass the current unified tracking model in many challenging scenarios.  
Specifically, under low-light conditions, the TIR modality typically outperforms the RGB modality. Since we did not prioritize the RGB modality but instead allowed the model to adaptively perceive and select the more effective modality, performance improvements were achieved. Moreover, the last two sequences in Fig.~\ref{fig:tracker_contrast} further validate this observation. In terms of speed, as shown in Tab.~\ref{tab:fps}, despite utilizing dual template frames, our model maintains a speed advantage by circumventing complex modal interactions.

\noindent \textbf{More visualization} To evaluate APTrack's generalization capability in RGB-D and RGB-E tasks, we conducted a visual analysis in four challenging scenarios: occlusion, similar object interference, cluttered backgrounds, and small targets. This is shown in Fig.~\ref{fig:contrast} from top to bottom. In the occlusion scenario, APTrack successfully captured the features of the occluded target by adaptively selecting the modality, thereby avoiding target loss, which is a common issue in traditional methods that rely on a fixed modality. In situations involving similar object interference, APTrack demonstrated its ability to flexibly choose the modality that best distinguishes the target from the interference, thereby enhancing target recognition and avoiding tracking errors. When faced with cluttered backgrounds, APTrack effectively separated the target from the background by adaptively selecting the modality that best highlights the target, ensuring accurate tracking in complex environments. For small targets, APTrack enhanced the perception of minute targets by integrating the strengths of multiple modalities. These results highlight APTrack's robustness and generalization ability in complex scenarios, attributed to its approach of adaptively selecting and integrating various modality features, rather than relying on a fixed dominant modality.

Additionally, we conducted a visual analysis of the attention maps for the search area to illustrate two challenging sequences. Fig.~\ref{fig:heat}~(a) shows that, under the RGB modality, the model's target imaging performance is superior compared to the TIR modality, while Fig.~\ref{fig:heat}~(b) indicates that the imaging performance is more prominent under the TIR modality. As shown in Fig.~\ref{fig:heat}, by not fixing any modality during the modeling process, APTrack can flexibly adjust its focus and fully leverage the complementary advantages of each modality, resulting in precise target attention in complex scenarios.

\section{Conclusions}

In this work, we investigated methods to enhance the performance of unified parameter trackers. Specifically, our proposed APTrack uniformly processes all modalities without additional optimization for specific ones, enabling the model to self-perceive and select modalities. Additionally, the designed AMI module effectively reduces redundancy and computational load, while the global modal preceptor facilitates modality alignment and advantage transmission. APTrack demonstrates outstanding performance across five benchmarks in three multi-modal tasks, showcasing its superior capabilities.

\textit{Limitation.} Adaptive perception has significant advantages: on the one hand, it eliminates the need to deliberately select the dominant-auxiliary modes, which makes the use of modes more flexible; on the other hand, it can effectively improve the performance of multi-modal tasks. However, there are some limitations in this adaptive perception approach, as it needs to give the same degree of processing to both modes, which will lead to an increase in the amount of computation to a certain extent, and put forward a higher demand for computational resources. Besides, in the natural language tracking task, natural language is different from image features and cannot be directly applied to the model we designed. In our follow-up work, we plan to use natural language as an independent prompt to assist the RGB-X tracking task, to further explore its potential application and optimize the tracking effect.

\end{document}